\documentclass[pdflatex,sn-nature]{sn-jnl}

\usepackage{graphicx}%
\usepackage{multirow}%
\usepackage{amsmath,amssymb,amsfonts}%
\usepackage{amsthm}%
\usepackage{mathrsfs}%
\usepackage[title]{appendix}%
\usepackage{xcolor}%
\usepackage{textcomp}%
\usepackage{manyfoot}%
\usepackage{booktabs}%
\usepackage{algorithm}%
\usepackage{algorithmicx}%
\usepackage{algpseudocode}%
\usepackage{listings}%

\theoremstyle{thmstyleone}%
\theoremstyle{thmstyletwo}%

\theoremstyle{thmstylethree}%

\begin{document}

\title[Article Title]{EMRModel: A Large Language Model for Extracting Medical Consultation Dialogues into Structured Medical Records}

\author[1]{\fnm{Shuguang} \sur{Zhao}}\email{shuguang6139@163.com}
\equalcont{These authors contributed equally to this work.}

\author[2]{\fnm{Qiangzhong} \sur{Feng}}\email{qzfeng@ustcinfo.com}
\equalcont{These authors contributed equally to this work.}

\author[3]{\fnm{Zhiyang} \sur{He}}\email{zyhe@iflytek.com}

\author[1]{\fnm{Peipei} \sur{Sun}}\email{404527175@qq.com}

\author[2]{\fnm{Yingying} \sur{Wang}}\email{wang.yingying@ustcinfo.com}

\author[3]{\fnm{Xiaodong} \sur{Tao}}\email{xdtao@iflytek.com}

\author[3]{\fnm{Xiaoliang} \sur{Lu}}\email{xllu@iflytek.com}

\author[3]{\fnm{Mei} \sur{Cheng}}\email{meicheng@iflytek.com}

\author[4]{\fnm{Xinyue} \sur{Wu}}\email{wuxinyue@usc.edu}

\author*[2]{\fnm{Yanyan} \sur{Wang}}\email{wang.yanyan@ustcinfo.com}

\author*[3,5]{\fnm{Wei} \sur{Liang}}\email{34316801@qq.com}

\affil[1]{\orgname{Taihe County People's Hospital}, \orgaddress{\city{Fuyang}, \postcode{236600}, \state{Anhui}, \country{China}}}

\affil*[2]{\orgdiv{Innovation + Research Institute}, \orgname{GuoChuang Cloud Technology Ltd.}, \orgaddress{\city{Hefei}, \postcode{230031}, \state{Anhui}, \country{China}}}

\affil[3]{\orgname{Xunfei Healthcare Technology Ltd}, \orgaddress{\city{Hefei}, \state{Anhui}, \postcode{230088}, \country{China}}}

\affil[4]{\orgname{University of Southern California}, \orgaddress{\city{Los Angeles}, \state{California}, \postcode{90089}, \country{USA}}}

\affil*[5]{\orgname{The First Affiliated Hospital of Anhui Medical University}, \orgaddress{\city{Hefei}, \state{Anhui}, \postcode{230022}, \country{China}}}


\abstract{Medical consultation dialogues contain critical clinical information, yet their unstructured nature hinders effective utilization in diagnosis and treatment. Traditional methods, relying on rule-based or shallow machine learning techniques, struggle to capture deep and implicit semantics. Recently, large pre-trained language models and Low-Rank Adaptation (LoRA), a lightweight fine-tuning method, have shown promise for structured information extraction. We propose \textbf{EMRModel},  a novel approach that integrates LoRA-based fine-tuning with code-style prompt design, aiming to efficiently convert medical consultation dialogues into structured electronic medical records (EMRs). Additionally, we construct a high-quality, realistically grounded dataset of medical consultation dialogues with detailed annotations. Furthermore, we introduce a fine-grained evaluation benchmark for medical consultation information extraction and provide a systematic evaluation methodology, advancing the optimization of medical natural language processing (NLP) models. Experimental results show EMRModel achieves an F1 score of \textbf{88.1\%}, improving by \textbf{49.5\% }over standard pre-trained models. Compared to traditional LoRA fine-tuning methods, our model shows superior performance, highlighting its effectiveness in structured medical record extraction tasks. }


\maketitle

\section{Introduction}\label{introduction}
\begin{figure}[ht]
    \centering
    \includegraphics[width=\linewidth]{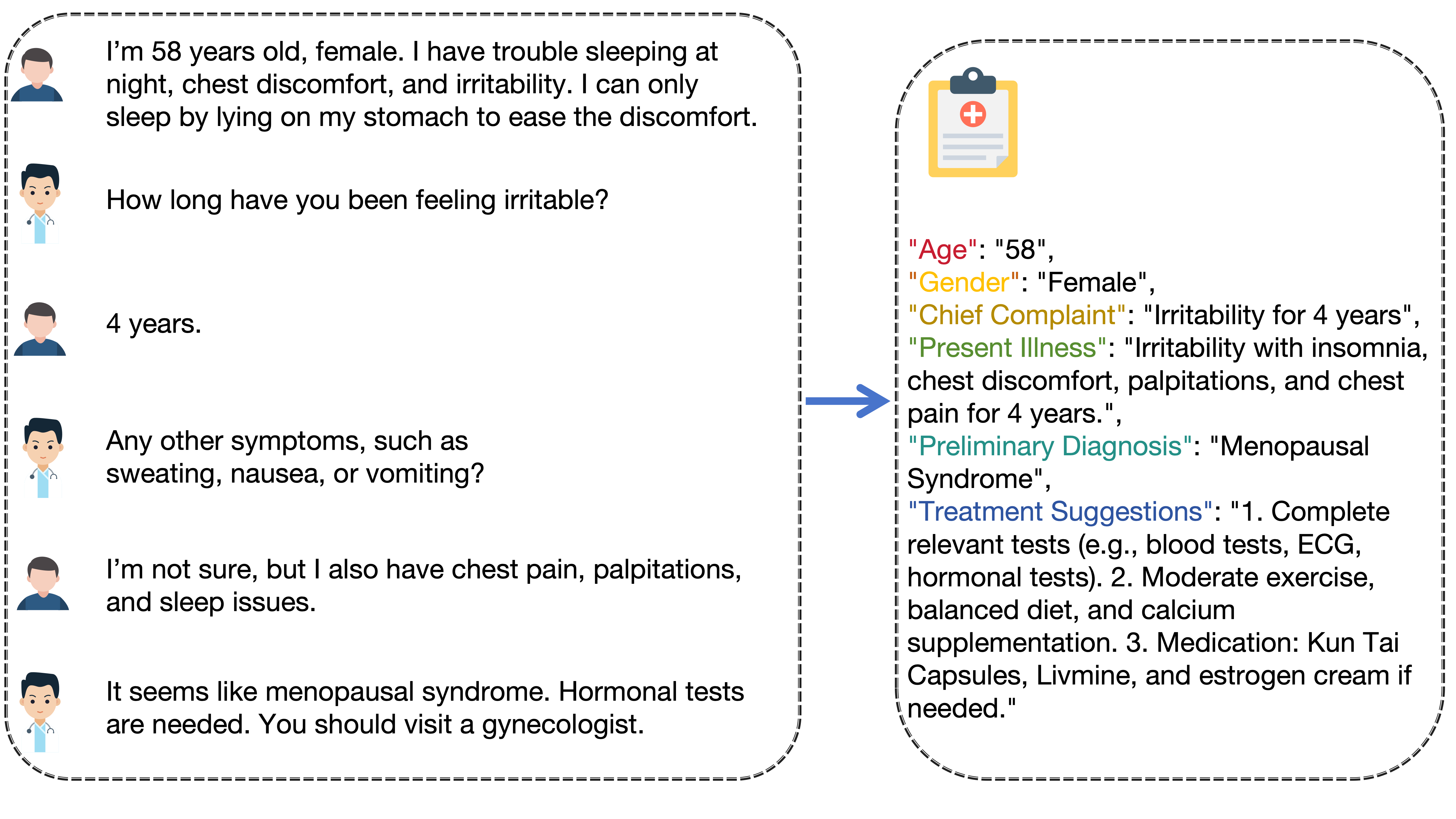} 
    \caption{\textbf{Example of converting medical consultation dialogues into structured data.}
}
    \label{fig:example}
\end{figure}

With the continuous advancement of healthcare informatization, digital recording of medical consultation dialogues has become an integral part of medical information systems. However, efficiently and accurately extracting relevant information to support subsequent diagnoses, treatments, and medical decision-making \citep{zhang2024rethinking} remains challenging. Figure \ref{fig:example} illustrates a typical unstructured doctor-patient consultation dialogue\citep{tan2024medchatzh}. Due to their unstructured nature, it has traditionally been difficult to directly apply these dialogues to downstream tasks, such as disease analysis \citep{pimenov2024state} and patient sentiment analysis \citep{eberhardt2025decoding}. By extracting and summarizing critical information \citep{xu2024extraction}, these unstructured dialogues can be transformed into structured key-value pairs, significantly benefiting future medical research.

Traditional methods of generating medical records typically rely on manual entry, rule-based template matching \citep{berge2023rulebased}, or deep learning techniques \citep{kang_using_2013, hernandez2022synthetic}, which encounter several limitations in practice. These include incomplete information, as existing automated methods often do not capture all critical details from consultations, thereby reducing the accuracy and usability of medical records. Additionally, non-standardized terminology used by different institutions or physicians poses challenges to standardizing information, complicating data integration and analysis. Furthermore, insufficient context understanding remains a significant issue, as current extraction methods often overlook the context-dependent nature of doctor-patient dialogues, failing to accurately interpret clarifications or corrections, ultimately affecting the coherence and precision of the resulting medical records.

In recent years, the rise of large language models (LLMs)  \citep{yang2022llm} has opened new opportunities for medical text processing. In particular, the combination of code-style prompt engineering \citep{white2024codeprompt} and LoRA (Low-Rank Adaptation) \citep{hu2022lora} fine-tuning techniques offers a novel solution to improve the efficiency and accuracy of medical information extraction. Unlike traditional fine-tuning methods, LoRA adapts low-rank matrices of LLMs \citep{xu2023lorallms}, significantly reducing computational costs while preserving the model's knowledge capacity, making it more suitable for specific tasks in the medical field. Additionally, code-style prompts provide structured guidance, enhancing the consistency and controllability of the model in medical record generation tasks.

Recent work \cite{mo_c-icl_2024} has shown that in the field of LLMs in-context learning (ICL) for information extraction tasks, code-style prompts can significantly improve performance. This is likely due to the fact that coder LLMs have learned a vast amount of structured information \citep{nam2024using}, and since information extraction tasks themselves are about structuring information, converting text into code-style prompts and inputting them into coder models yields better results. However, the paper also points out that using code-style prompts as ICL examples for information extraction tasks still falls significantly short of the performance achieved by supervised fine-tuning. Building on these insights, we consider applying code-style prompts in the fine-tuning stage for medical diagnostic information, where the output from the fine-tuned model in code format is then decoded into structured data.

We propose EMRModel, a novel method for extracting structured medical records from consultation dialogues. In this model, the input text is first processed by the Prompt-Encoder, which converts natural language instructions and consultation dialogues into code-style prompts. The transformed semi-structured data is then fed into a LLM for LoRA fine-tuning, where most of the model parameters are frozen and only the low-rank matrices are trained. The model captures latent relationships in the consultation dialogues through self-attention mechanisms and automatically extracts structured features. Finally, the Prompt-Decoder  converts the code-style output into structured results. Additionally, we propose a fine-grained evaluation benchmark for medical consultation information extraction and provide a systematic evaluation methodology, promoting the optimization of medical natural language processing (NLP) models.

The core contributions of this study are as follows:
\begin{itemize}
    \item \textbf{High-Quality Medical Dataset Construction.}
 We meticulously constructed a comprehensive medical consultation record extraction dataset, covering various diseases, consultation scenarios, and medical terminology, to meet the high-quality, domain-specific training and evaluation needs. This dataset not only enhances the training effectiveness of the model but also provides a standardized evaluation environment for future research.
    \item \textbf{A Novel Medical Electronic Record Extraction Model Based on Coding Style.}
We propose a new method combining LoRA fine-tuning with a coder-style prompt, making LLMs more efficient and accurate for the structured  electronic record extraction task in medical consultation dialogues. This approach effectively leverages the intrinsic capabilities of large language models while specifically fine-tuning them to address the complexity and professional nature of medical texts, thereby enhancing the model's performance in medical record summarization tasks.
    \item \textbf{Establishment of a Medical Consultation Record Extraction Benchmark.}
We developed a robust medical consultation dialogue record extraction benchmark, providing systematic evaluation for different information fields, offering reproducible comparative standards for future medical NLP research. This benchmark framework supports multi-field, fine-grained evaluation of medical information extraction, providing reference directions for the improvement and optimization of medical NLP models.
\end{itemize}

\section{Related Work}
\label{Related Work}

\subsection{Large Pretrained Language Models and Fine-Tuning Methods}

In recent years, LLMs \citep{vaswani_attention_2023,devlin_bert_2019} have demonstrated powerful capabilities across various fields of NLP \citep{noauthor_fine-tuning_nodate}. To adapt these models to specific domain tasks, fine-tuning techniques have been widely applied. These models have shown exceptional performance across various tasks, including text classification, question-answering systems, and machine translation \cite{raffel_exploring_2023}.

Compared to full-parameter fine-tuning, efficient fine-tuning techniques significantly reduce training time and costs while yielding similar fine-tuning performance \cite{ding_parameter-efficient_2023}. LoRA fine-tuning, as a novel efficient fine-tuning technique, reduces the fine-tuning cost by applying low-rank decomposition to some model parameters. It can rapidly capture domain-specific features while preserving the original model's semantic capabilities \cite{hu_lora_2021}.

\subsection{Medical Information Extraction and Structured Medical Record Generation}

Medical information extraction aims to extract valuable clinical information, such as medical history, diagnoses, and treatment plans, from unstructured medical texts. This task holds significant importance in the healthcare field as it assists doctors in quickly identifying key data from complex information, thereby improving diagnostic efficiency and accuracy. With the rise of online consultations, research in this area has been on the rise \cite{noauthor_clinical_2018}.

Traditional methods primarily rely on rule-based or statistical models, which often require manually written rules or the use of statistical models for information extraction \citep{kang_using_2013} . Although these methods can be effective in specific scenarios, they face limitations due to their strong dependence on rules, restricted scope, and poor adaptability to new data. As a result, they encounter many challenges in practical applications. In recent years, deep learning-based methods have made significant progress in this field, particularly with the development of NLP technologies, enabling deep learning models to better understand the language of the medical field and automatically extract key medical information from unstructured data \cite{deshmukh_information_2021}. However, accurately capturing and structuring medical record information from complex consultation dialogues remains a challenging research problem.

\subsection{Prompt Design}

Prompt techniques play a crucial role in guiding large models to perform tasks effectively. A well-designed prompt not only enhances the model's efficiency but also significantly improves its performance on specific tasks. With the release of coder large models, using code-style prompts to guide models through related tasks has become a growing trend. For instance, CodeIE \cite{li_codeie_2023} proposed using code-style prompts for context-based question answering, and the results demonstrated that code-style prompts outperformed directly using natural language prompts in certain scenarios \cite{nie_code-style_2023}. Similarly, CodeKGC \cite{bi_codekgc_2024} showcased the significant impact of code-style prompts in knowledge graph construction, indicating that this method is more effective in guiding the model for generative tasks.

Moreover, in text classification tasks, code-style prompts also exhibited excellent performance, enhancing large language models' performance on this task \cite{mohajeri_cocop_2024}. In the field of information extraction, C-ICL \cite{mo_c-icl_2024} were the first to apply code-style prompts in large models' ICL tasks, achieving remarkable results. The research shows that using code-style prompts for ICL information extraction tasks on coder models yields higher accuracy \cite{li_codeie_2023}. However, despite the promising results of using code-style examples in ICL tasks, previous studies have also shown that the overall performance boost from code examples still cannot surpass the gains achieved through supervised fine-tuning.

Building on this foundation, this study further explores the possibility of using code-style prompts combined with LoRA fine-tuning on large natural language processing models. We found that LoRA fine-tuning, when combined with code-style prompts, significantly improves performance, especially in information extraction tasks, demonstrating the effectiveness of this approach in enhancing model performance.

\section{Materials and methods}
\label{Materials and methods}

\begin{figure}[ht]
    \centering
    \includegraphics[width=\linewidth]{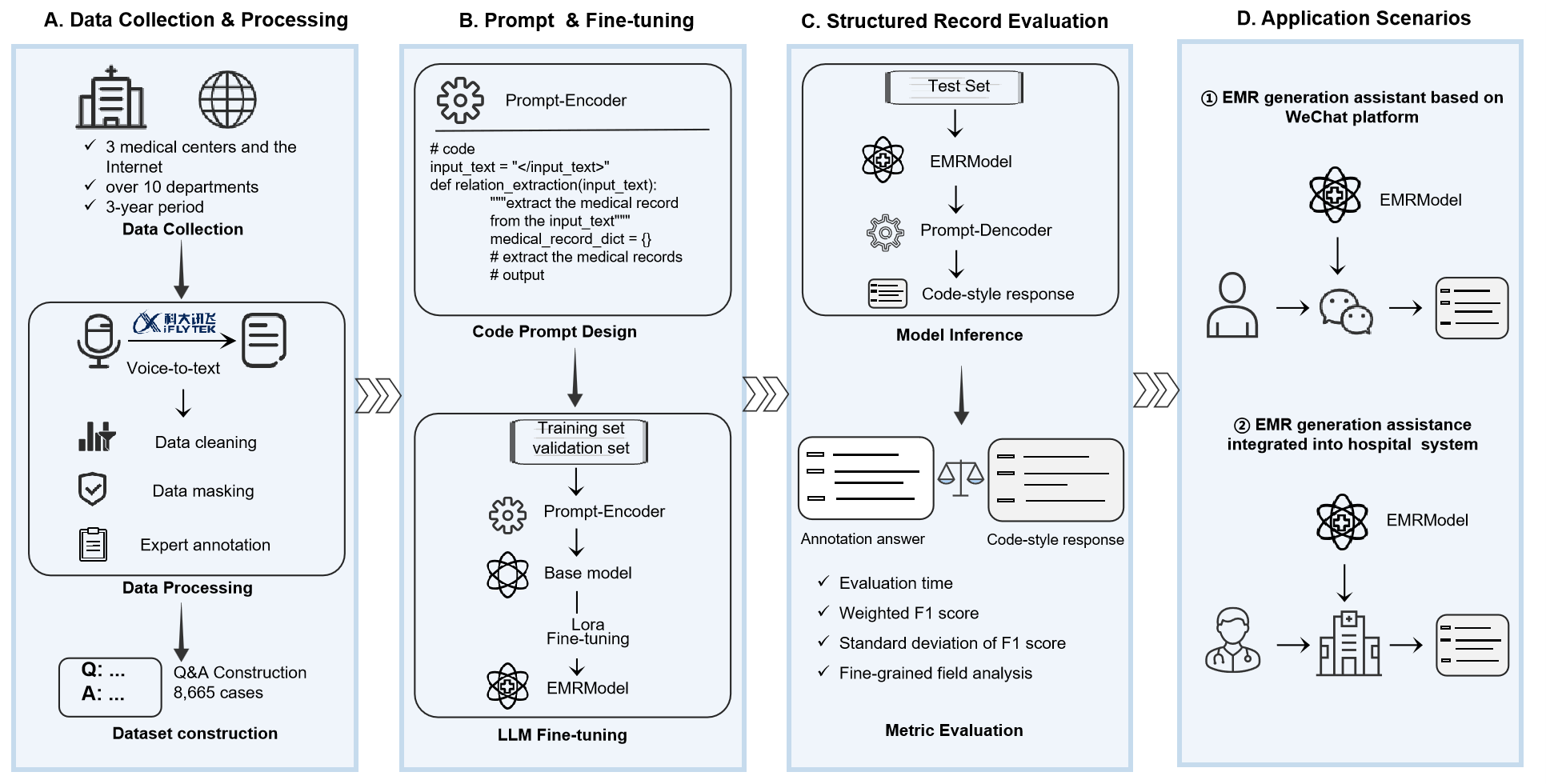} 
    \caption{Study design overview.
}
    \label{fig:model}
\end{figure}

This study was approved by the Ethics Committee of the First Affiliated Hospital of Anhui Medical University and Taihe County People's Hospital. As illustrated in Figure \ref{fig:model}, the research comprised the following four phases: (1) We collect physician-patient dialogue audio from over 10 departments across two medical centers. Following audio-to-text conversion, data cleaning, de-identification, and expert annotation, we construct a dataset comprising 8,665 physician-patient encounter records. (2) We design two types of prompt templates, Natural Language (NL) and Code, and then efficiently fine-tune the LLMs using the LoRA technique with these prompts. (3) Utilizing independent test data allows for evaluating the EMR generation capabilities of different LLMs under the various prompting strategies from multiple dimensions. (4) The best-performing model is subsequently deployed as two practical tools: a WeChat platform-based EMR generation assistant and an EMR generation support tool integrated into the hospital's information system.

\subsection{Task Definition }
The goal of this study is to convert medical consultation dialogues into structured medical records, covering content such as patient demographics, chief complaints, medical history, past medical history, and treatment recommendations.

\subsection{Dataset Construction}

This study focuses on the automatic generation of EMR within the medical domain, constructing a dataset based on authentic medical consultation records. Prior to utilization, all data underwent rigorous de-identification to ensure patient privacy. The preprocessing work primarily involved text cleaning, segmentation, and the annotation of key information, aimed at providing standardized input for subsequent model training.

\subsubsection{Raw Data Collection and Preprocessing}

The data for this study originate from authentic physician-patient consultation dialogues from over ten departments across three medical centers. To ensure the completeness and clarity of data collection, professional recording devices featuring high-sensitivity pickup and long-duration recording capabilities are deployed in multiple outpatient consultation rooms to capture the dialogue audio throughout the entire consultation process. In total, the collected raw data comprises approximately 46,910 consultation records. After obtaining the raw audio, the primary step is to convert the audio data into text format using iFLYTEK's speech-to-text technology, integrated with a custom-built medical lexicon. The initially obtained set of physician-patient dialogue texts is denoted as:
\begin{equation}
D_{\mathrm{raw}} = \left\{ x_i \right\}_{i=1}^N,
\end{equation}
where $x_i$ represents the $i$-th dialogue text, and $N$ is the total number of dialogue texts.

Subsequently, a series of preprocessing operations, represented by the function $\mathcal{F}$, are performed on the raw data $D_{\mathrm{raw}}$. The specific steps include: (1) Text Cleaning: Aiming to remove noise such as text transcribed from irrelevant background noise and colloquial filler words. (2) Syntactic and Semantic Normalization: Adjusting sentence structures and standardizing the expression of medical terminology. (3) Correction of Raw Data Anomalies: Addressing issues like garbled text or unintelligible segments resulting from speech recognition. This involves manual review by listening to the original audio for correction, utilizing contextual information and medical knowledge to rectify misrecognized medical terms. (4) Data Security and Privacy Protection: Strictly adhering to medical data privacy regulations, all sensitive personal information appearing in the text, such as patient names, ID numbers, addresses, and contact information, is thoroughly de-identified and masked. Following these preprocessing steps, we obtain 8,665 standardized physician-patient dialogue records:
\begin{equation}
D = F(D_{\mathrm{raw}}) = \left\{ F(x_i) \right\}_{i=1}^M,
\end{equation}

Finally, for each processed dialogue record $F(x_i)$, structured medical records are generated using a semi-automatic annotation method. This approach combines preliminary extraction by NLP tools with manual review and correction by professional annotators. Assuming each structured medical record consists of multiple sections (e.g., patient information, chief complaint, history of present illness, etc.), let $\mathcal{F}$ denote the set of fields corresponding to these sections. Then, the structured medical record $y_i$ corresponding to the $i$-th dialogue can be represented as:
\begin{equation}
y_i = \{ y(i, f) \mid f \in \mathcal{F} \}.
\end{equation}
where $y(i, f)$ represents the value for field $f$ in the $i$-th sample.

\subsubsection{Construction of the Question-Answering Pair }

Upon completing the preprocessing of the physician-patient dialogue texts and structured medical records, we further construct a Question-Answering ($Q\&A$) pair dataset specifically tailored for the task of automatic EMR generation. In this process, we treat each physician-patient dialogue text as the input question ($Q$) and the corresponding structured medical record as the target output answer ($A$). This is formally represented as:
\begin{equation}
D_{QA} = \left\{ (F(x_i), y_i) \right\}_{i=1}^N
\end{equation}
where $F(x_i)$ represents the $i$ physician-patient dialogue text, and $y_i$ represents the corresponding structured medical record. Each pair $((F(x_i), y_i)$ constitutes a complete sample for training or evaluation.

\subsection{LLM Fine-tuning Strategies for EMRs Generation}

To enable LLMs to accurately understand physician-patient dialogues and generate standardized EMR, we propose the EMRModel method. This method designs specific code-style prompt templates, transforms the EMR generation task into a structured code completion problem, and employs the parameter-efficient LoRA fine-tuning technique. This approach allows the model to enhance its professional performance in the medical domain while retaining its general capabilities.

\subsubsection{Design of Code-Style Prompts}
\begin{figure}[ht]
    \centering
    \includegraphics[width=\linewidth]{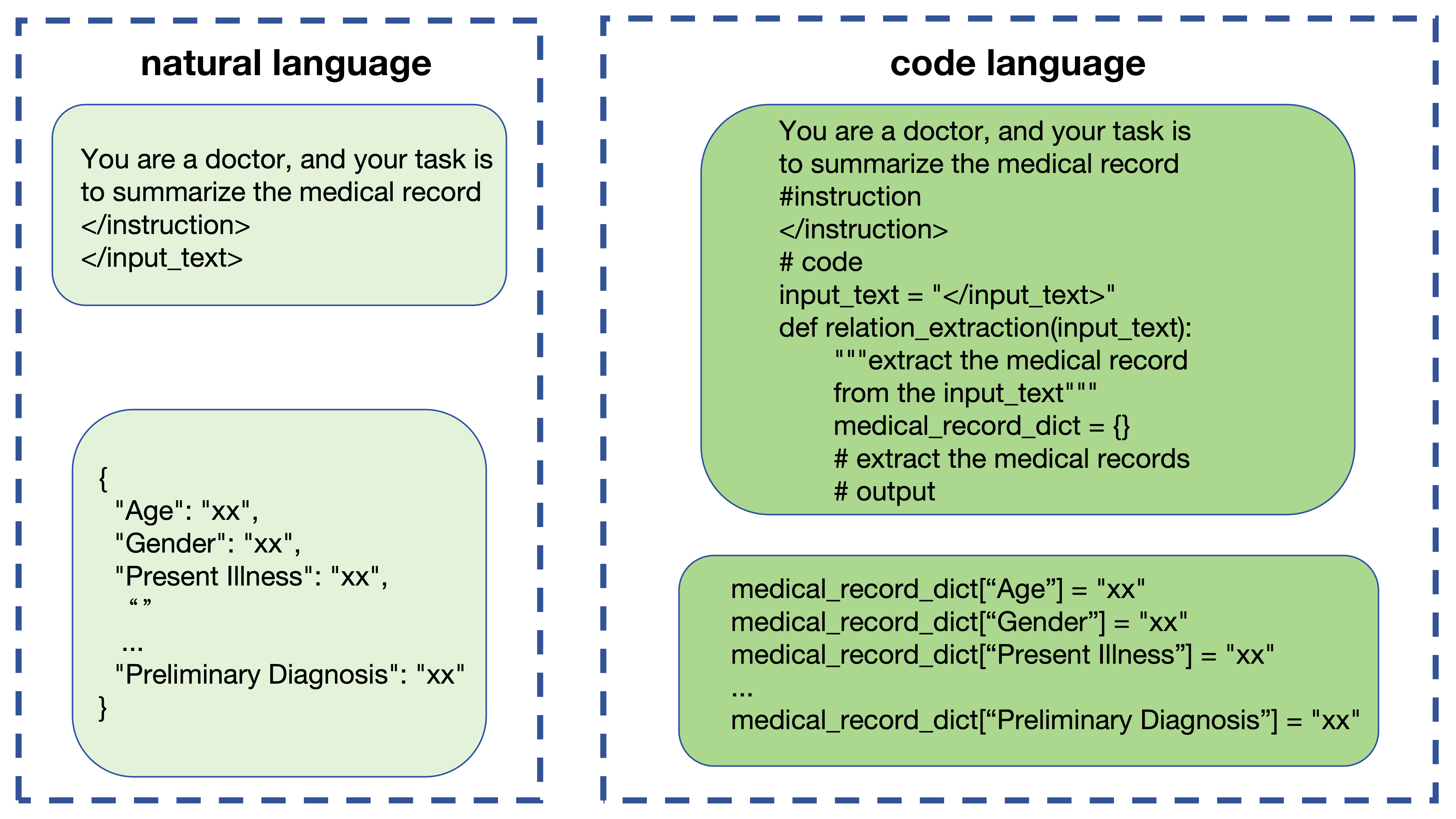} 
    \caption{\textbf{A comparison diagram between the traditional natural language prompt structure (left) and the code-style prompt structure (right).}
}
    \label{fig:nl_vs_code.png}
\end{figure}

To enhance the output precision and standardization of LLMs in the structured EMRs generation task, this study proposes and implements a code-style prompt strategy. To support the implementation of this strategy, we construct a prompt-encoder module. Its core function is the automated batch encoding transformation of input physician-patient dialogue texts using a predefined structured code template $T$. Specifically, this module embeds the processed dialogue text $F(x_i)$ into a specific code framework and achieves semantic constraint by integrating task instructions. This encoding process can be formally represented as:
\begin{equation}
p_i = E(F(x_i), \mathcal{T})
\end{equation}
where $E$ represents the transformation function executed by the prompt-encoder, ultimately generating the code-style prompt $p_i$ for use by the LLM. Notably, employing code-style prompts alters the model's expected output paradigm. It guides the LLM away from generating data structures like natural language text, focusing it instead on completing specific fields within the input code snippet. This transforms the EMRs generation task into a controlled code completion problem. As shown in Figure \ref{fig:nl_vs_code.png}, compared with traditional natural language prompts, this method significantly enhances the model's ability to constrain the output format through explicit code syntax, effectively reducing the impact of unstructured text noise on the results.

\subsubsection{LLM Fine-tuning based on LoRA}

To enable the pre-trained LLM to effectively understand physician-patient dialogues and generate standardized EMRs, this study employs the LoRA technique for domain-specific adaptation and fine-tuning within the medical field. The specific steps include: (1) Parameter Freezing and Update: By freezing the majority of the pre-trained model's parameters and fine-tuning only the low-rank parameters, this approach preserves the model's original language understanding capabilities while enhancing its sensitivity to medical domain text. (2) Low-Rank Adaptation Layers: Introducing low-rank adaptation layers into the model enables it to more effectively capture the semantic features specific to the medical domain. (3) Training Strategy: Through careful setting of the learning rate and the number of training steps, the strategy ensures efficient convergence of the model on medical text processing tasks and improves its adaptability. Specifically, Let the pre-trained large model be denoted as $M$ with parameter set $\theta$. In the LoRA method, for the weight matrix $W \in \mathbb{R}^{d \times d}$ that needs to be fine-tuned in the model, we introduce a low-rank update term. Specifically, the weight update is expressed as:
\begin{equation}
W^\ast=W+\mathrm{\Delta W},\mathrm{\Delta W}=BA,
\end{equation}
where $A \in \mathbb{R}^{r \times k}$ and $B \in \mathbb{R}^{d \times r}$, with $r$ being the low-rank dimension. During fine-tuning, the pre-trained parameters $\theta$ remain unchanged, and only the LoRA parameters $A$ and $B$ are optimized. The fine-tuning objective function of the model is:
\begin{equation}
\min_{A, B} \sum_{i=1}^{|D'|} \mathcal{L}\left(M(p_i; \theta, A, B), y_i\right),
\end{equation}
where $\mathcal{L}(\cdot, \cdot)$ represents the cross-entropy loss function, which measures the discrepancy between the structured medical records generated by the model and the true annotations. After fine-tuning, the output from the LLM remains semi-structured code-style content. Finally, this output is fed into a prompt-decoder module for parsing, which then batch-converts it into the required structured EMR information.

\section{Experiments}

\subsection{Experimental setup}

\subsubsection{Experimental Dataset}

\begin{figure}[ht]
    \centering
    \includegraphics[width=\linewidth]{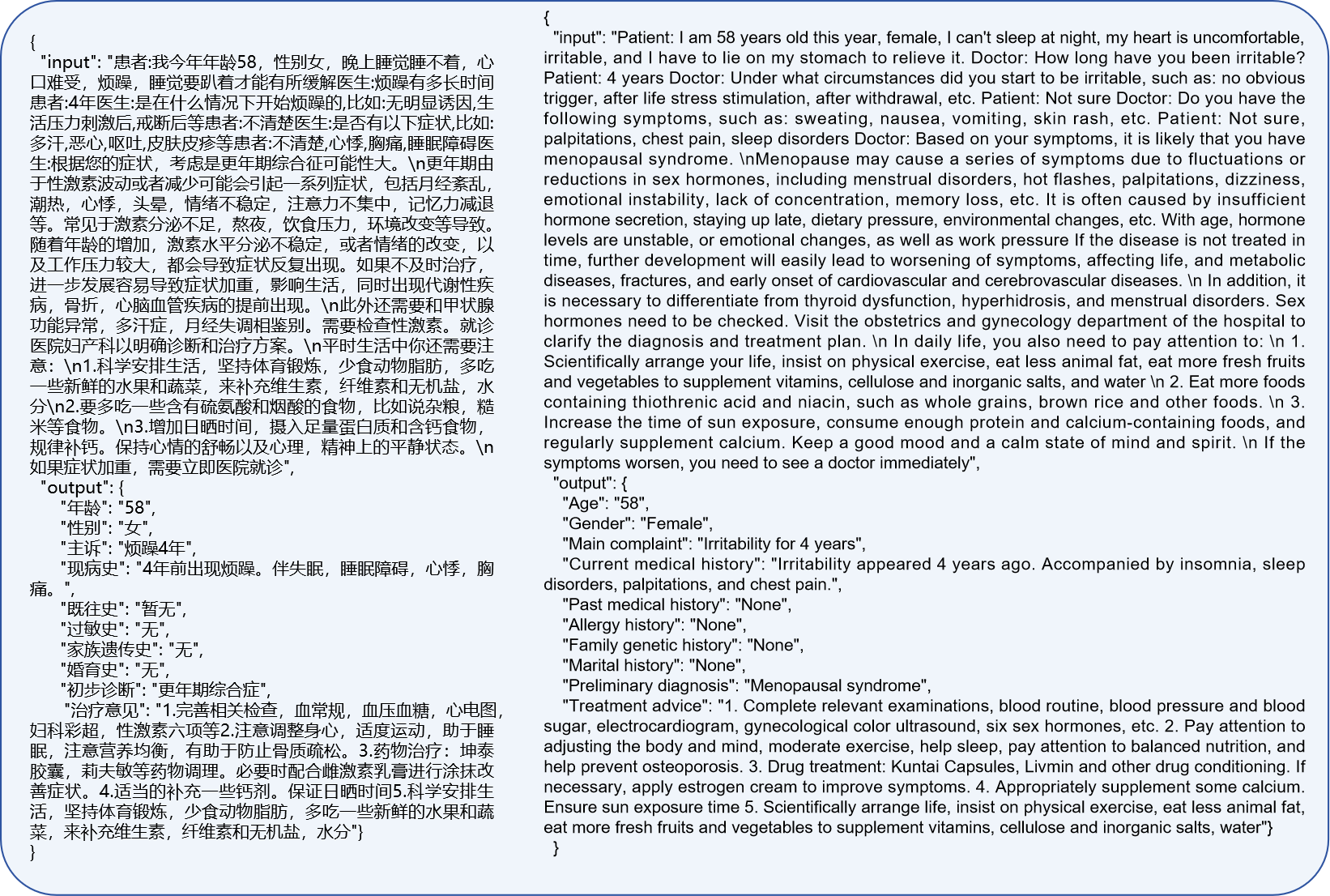} 
    \caption{\textbf{Dataset data example.}
}
    \label{fig:data_example}
\end{figure}

Our data was primarily sourced from the First Affiliated Hospital of Anhui Medical University, Taihe County People's Hospital, Beijing 301 Hospital, and online sources. Professional recording devices were deployed in multiple outpatient consultation rooms across various departments—including Geriatrics, Gastroenterology, Thyroid and Breast Surgery, Neurology, Cardiology, and Respiratory Medicine—to capture audio recordings of physician-patient interactions during diagnosis and treatment. Data collection occurred over a 3-year period from April 2021 to April 2024, comprehensively covering consultations across different seasons, weekdays, and weekends to ensure broad data representativeness. From this effort, approximately 46,910 medical records were initially collected. After processing and cleaning, 8,665 records were retained for the study. These were subsequently divided into training, validation, and test sets, comprising 7,329, 800, and 536 records, respectively. The specific data example is shown in Figure \ref{fig:data_example}.

\subsubsection{Evaluation Metrics}

To comprehensively and accurately evaluate the performance of the LLM on the task of generating structured medical records from unstructured text, this study employs the weighted average F1 score as the primary evaluation metric. To assess the model's performance stability across different samples, we further calculate the sample standard deviation of the weighted average F1 scores obtained for each individual sample. This standard deviation reflects the variability in the model's performance when processing different test samples. A lower standard deviation indicates more stable model performance.

The weighted average F1 score is designed to provide a consolidated measure of the LLM's precision and recall in information extraction, weighted according to the importance of different fields. Its calculation process is as follows: First, for each target field $f$ in the medical record template, the F1 score for that field, denoted as $\text{F1}_f$, is calculated based on precision and recall. During this process, if a field is missing in the LLM's output, it is treated as empty. Next, the weight $w_f$ for each field is determined, defined as the number of characters in that field within the ground truth, thereby reflecting the relative importance of different fields within the overall medical record. Finally, the weighted average of the F1 scores across all fields is calculated using the following formula to obtain the overall F1 score:
\begin{equation}
\text{F1}_{\text{overall}} =
\begin{cases}
    \dfrac{\sum_{f \in \mathcal{F}} w_f \cdot \text{F1}_f}{\sum_{f \in \mathcal{F}} w_f}, & \text{if } \sum_{f \in \mathcal{F}} w_f > 0 \\
    0, & \text{otherwise}
\end{cases}
\end{equation}
where $\mathcal{F}$ represents the set of all target fields. This weighted averaging method effectively reflects the model's comprehensive performance in extracting various key information components, avoiding potential evaluation biases that might arise from a simple arithmetic mean.

\subsection{Experimental Results}

This study aims to systematically investigate the impact of different combination strategies involving LLM types (Natural Language LLMs vs. Coder LLMs) and prompt formats (Natural Language format vs. Code language format), along with the LoRA fine-tuning technique, on the performance of structured medical information extraction. To achieve this goal, we designed and compared eight different methods and models.

Among these, Four methods leverage the LoRA fine-tuning technique to elucidate the effects of different combination strategies. We selected the Qwen2.5-7B series models as the base models. The specific fine-tuning strategies include:
\begin{itemize}
    \item \textbf{Coder Model + NL Prompt:} The code-specialized LLM Qwen2.5-Coder-7B-Instruct is fine-tuned using data formatted in natural language.
    \item \textbf{Coder Model + Code Prompt:} The Qwen2.5-Coder-7B-Instruct model is fine-tuned using data in a code-style format.
    \item \textbf{NL Model + NL Prompt:} The general-purpose natural language model Qwen2.5-7B-Instruct is fine-tuned with natural language-formatted data.
    \item \textbf{NL Model + Code Prompt (our model):} The Qwen2.5-7B-Instruct model is fine-tuned using data in code-style format. This configuration represents our proposed approach. 
\end{itemize}

Furthermore, to quantify the specific gains provided by fine-tuning and to establish performance benchmarks, we also evaluated four baseline LLMs without any task-specific fine-tuning:
\begin{itemize}
    \item \textbf{Spark-medical-X1:} A commercial LLM developed by iFLYTEK, specifically optimized for the medical domain to provide expert-level medical information processing capabilities.
    \item \textbf{DeepSeek-V3:} A general-purpose LLM developed by DeepSeek, built upon a Mixture of Experts (MoE) architecture and primarily designed for natural language processing tasks.
    \item \textbf{DeepSeek-R1-671B:} A large-scale reasoning model by DeepSeek, tailored for complex reasoning tasks, with enhanced capabilities in mathematics, code generation, and logical inference.
    \item \textbf{DeepSeek-R1-32B:} A dense model distilled from DeepSeek-R1-671B based on the Qwen model architecture, designed to strike a balance between high performance and computational efficiency. 
\end{itemize}

\begin{figure}[ht]
    \centering
    \includegraphics[width=\linewidth]{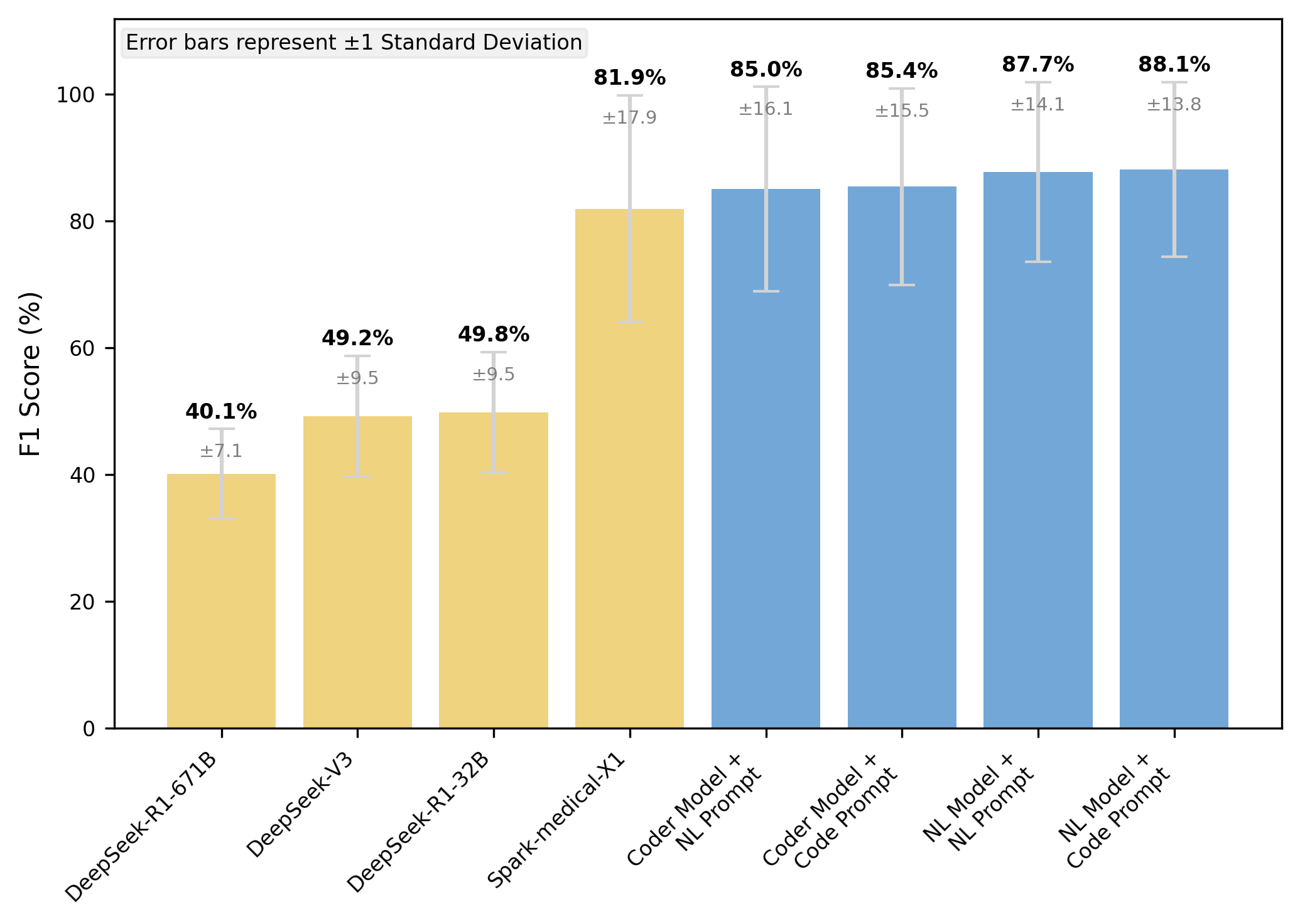} 
    \caption{Performance comparison of different models and fine-tuning strategies in the task of extracting information from medical consultation dialogues.
}
    \label{fig:all_model_comparison}
\end{figure}

Figure \ref{fig:all_model_comparison} illustrates the F1 scores and the sample standard deviation of F1 scores for the different methods on the medical dialogue information extraction task. Overall, LoRA fine-tuning significantly enhanced model performance, with all fine-tuned strategies outperforming the non-fine-tuned baseline models. Among these, the “NL Model + Code Prompt” strategy performed best, achieving the highest F1 score of 88.1\%, which preliminarily verifies the effectiveness of this combination strategy.

Evaluation of the baseline models shows that Spark-medical-X1, which has been optimized for the medical domain, achieved a relatively strong F1 score of 81.9\% without any fine-tuning. However, it still underperformed compared to all fine-tuned models. This is primarily because our fine-tuning was tailored to the specific task dataset, whereas Spark-medical-X1, being a general-purpose medical model, though equipped with domain knowledge, lacked task-specific training. In contrast, the DeepSeek series of general-purpose models, without any fine-tuning, performed significantly worse, with F1 scores ranging from 40.1\% to 49.8\%. This indicates that these powerful general-purpose models, without task-specific fine-tuning, struggle to effectively adapt to and perform structured information extraction tasks in such specialized domains directly. Notably, despite the poor F1 performance of the DeepSeek models, their sample standard deviation for F1 scores was relatively low. This might reflect a certain consistency in the models' output structure or numerical handling, but it did not translate into effective key information extraction capabilities.

Among the four LoRA fine-tuning strategies, the fine-tuned NL Models generally outperformed the Coder Models, and Code-style prompts provided gains for both model types to varying degrees. Specifically, the advantage of the NL Model lies in its superior semantic understanding capabilities, crucial for grasping the nuances of complex medical dialogues. Its peak F1 score reached 88.1\%, significantly exceeding the Coder Model's 85.4\%. The Code Prompt enhanced performance for both model types by providing structured guidance. It helped the NL Model accurately translate semantic understanding into structured outputs, boosting its F1 score from 87.7\% to 88.1\%. Simultaneously, it aligned better with the Coder Model's inherent strengths in handling structured data, increasing its F1 score from 85.0\% to 85.4\%.

In summary, fine-tuned NL Models are better suited for this task than Coder Models, and introducing well-designed Code-style prompts as structured guidance can further enhance model performance.

\subsection{Sensitivity analysis}
This section provides a detailed analysis of the experimental results, evaluating the performance of EMRModel under different experimental settings. We primarily focus on metrics such as F1-score and Standard Deviation of F1 Score, and by comparing various methods, we explore the role of LoRA fine-tuning and coder-style prompts in medical consultation information extraction.

\subsubsection{Base Model Selection}

\begin{figure}[ht]
    \centering
    \includegraphics[width=\linewidth]{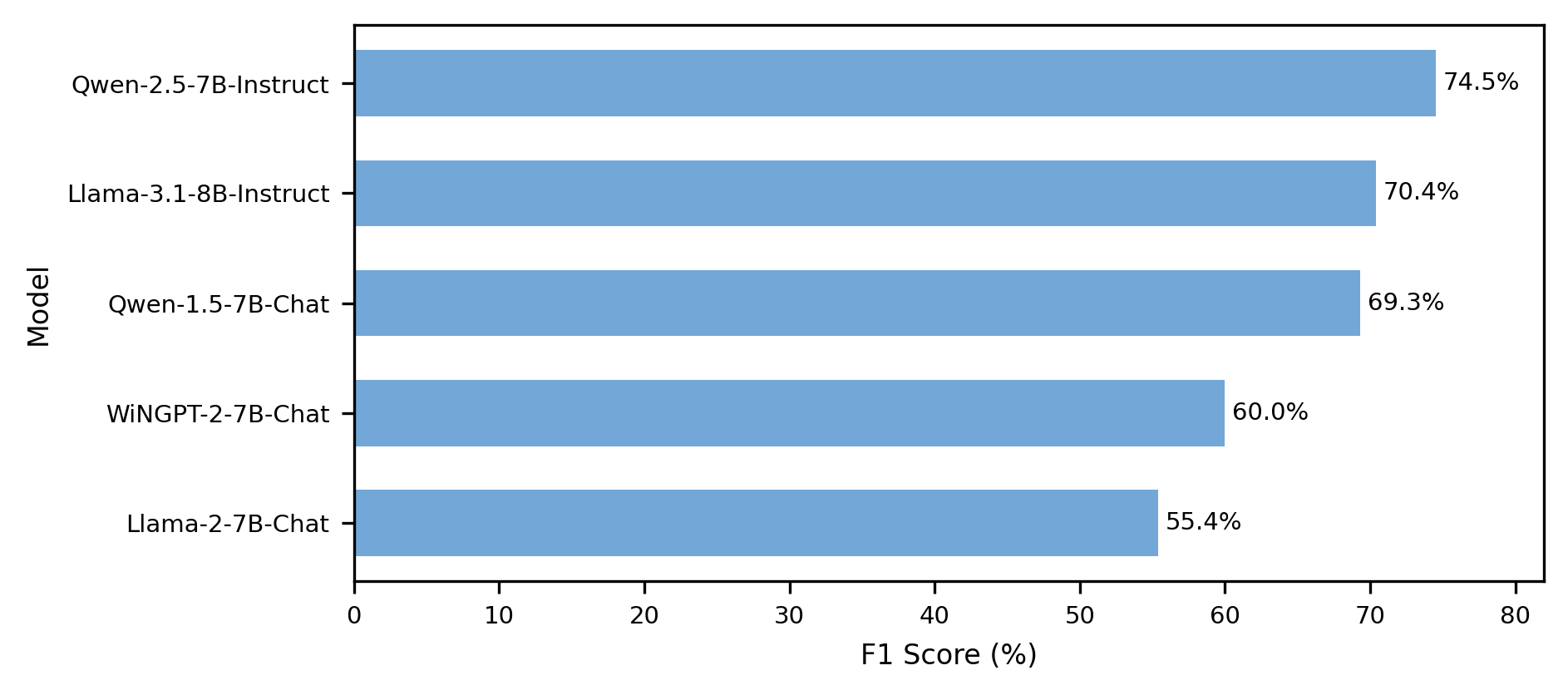} 
    \caption{\textbf{Scores of different models on LoRA fine-tuning in one epoch.}
}
    \label{fig:base_model_selection}
\end{figure}

Considering factors such as training cost, performance, and whether the model is open-source and fine-tunable, we selected Llama-2-7B-Chat, Llama-3.1-8B-Instruct, Qwen-1.5-7B-Chat, and Qwen-2.5-7B-Instruct. These models were fine-tuned using LoRA for one epoch to evaluate their performance on our medical data, which helped us in selecting the most suitable base model. Given the presence of many medical terms in the text, we also considered the open-source medical large model WiNGPT-2-7B-Chat. After fine-tuning and testing the F1 scores of the above models, the results are shown in Figure \ref{fig:base_model_selection}. After considering various factors, we ultimately chose Qwen-2.5-7B-Instruct as our fine-tuning base model.

\subsubsection{Analysis of base model code style and natural language style results}

\begin{figure}[ht]
    \centering
    \includegraphics[width=0.5\textwidth]{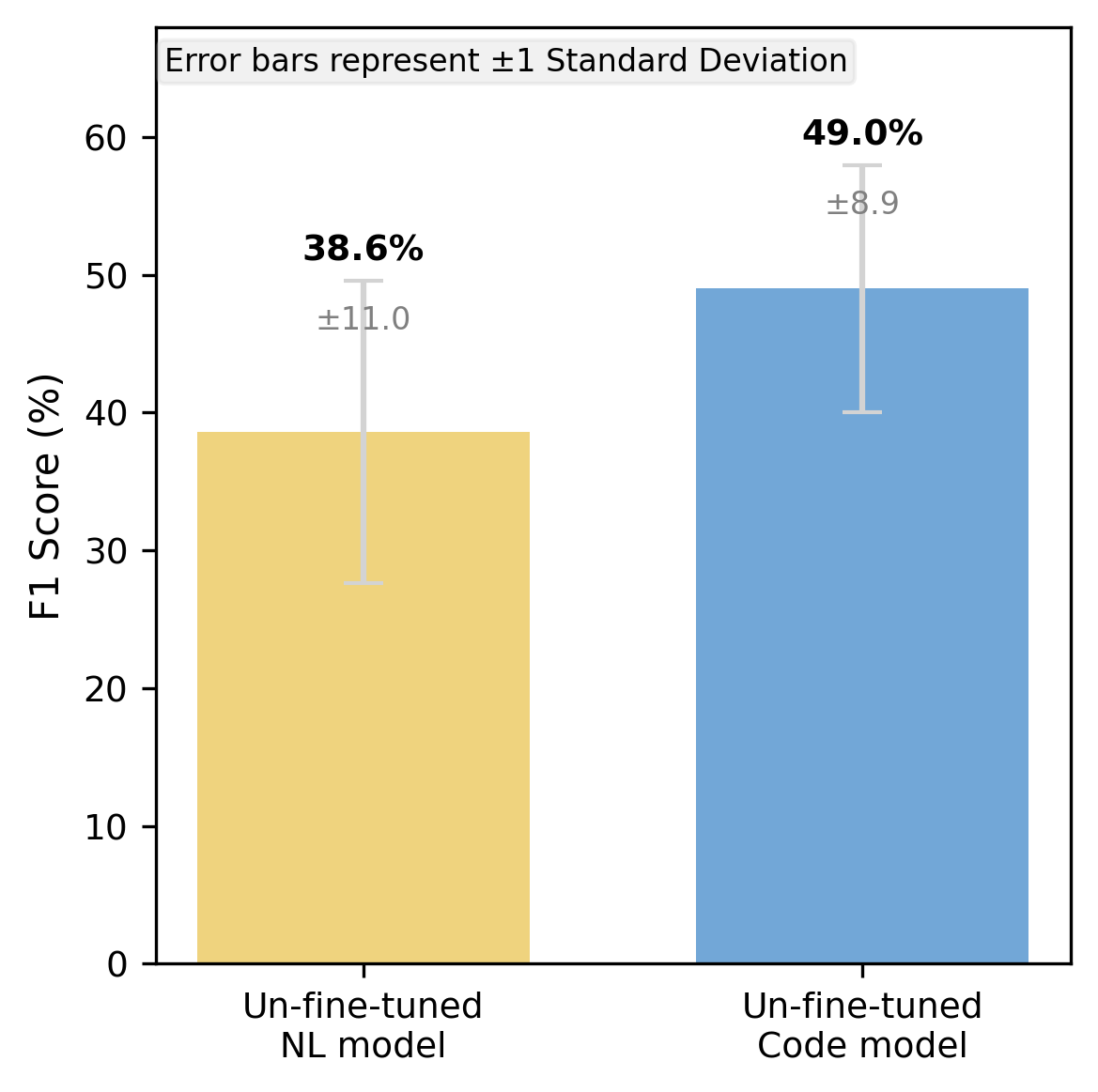} 
    \caption{\textbf{The scores of the Qwen-2.5-7B natural language model and the coder model on the  medical summarization task before fine-tuning.}
}
    \label{fig:before_fine_tuning}
\end{figure}

Although traditional coder models demonstrate strong structured data processing capabilities in information extraction tasks using ICL approaches, directly applying them in the fine-tuning process of natural language large models has not fully realized their potential advantages. To compare the performance of the Coder model and the NL model on our task before fine-tuning, we conducted a comparative experiment, and the results are shown in Figure \ref{fig:before_fine_tuning}. As observed, before fine-tuning, the Coder model, when combined with structured prompts, was more suitable for information extraction tasks. Models trained using structured methods perform better in information extraction tasks under in-context learning approaches. This advantage is likely due to the Coder model's extensive training on structured data during the pre-training phase, which gives it a natural advantage in understanding and generating structured information. In contrast, general-purpose NL models primarily focus on open-domain natural language generation and understanding and lack optimization specifically for structured information extraction. Therefore, without fine-tuning, the NL model's ability to understand structured features in information extraction tasks is weaker, resulting in suboptimal performance compared to the Coder model.

\subsubsection{ Fine-Grained Field Extraction Analysis}

\begin{figure}[ht]
    \centering
    \includegraphics[width=\linewidth]{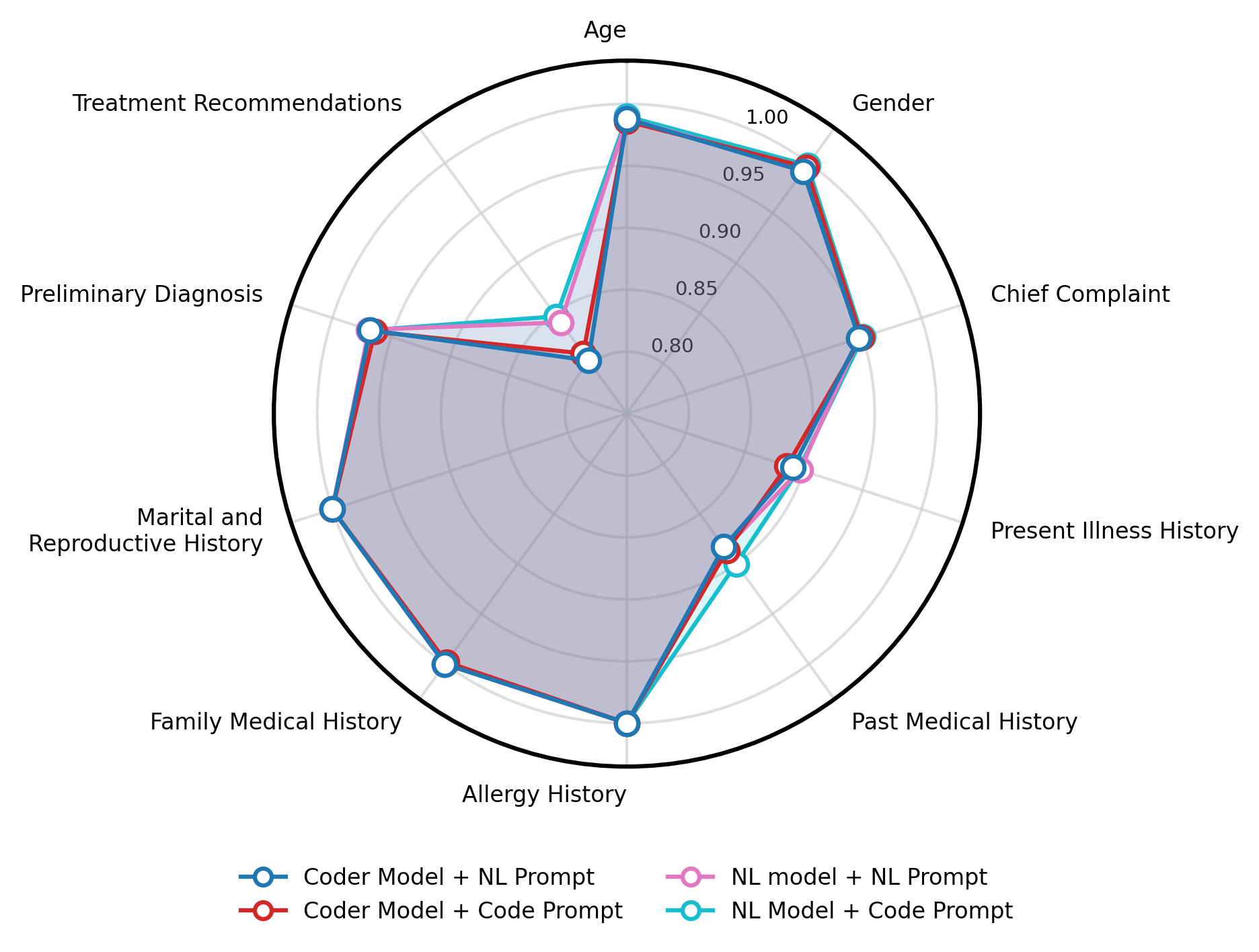} 
    \caption{\textbf{The F1-scores of different model combinations across various fields.}
}
    \label{fig:fine_grained_analysis}
\end{figure}

We conduct a fine-grained analysis of the F1 scores achieved by different model combinations across various fields to evaluate the role of prompt design and the LoRA fine-tuning technique in different information extraction sub-tasks. The results are presented in Figure \ref{fig:fine_grained_analysis}. the experimental results show that for highly structured fields with relatively fixed formats, such as age, gender, family medical history, maritaland reproductive history, and allergy history, the F1 scores for all model combinations approach 1.0. This indicates that the consistent structure of these fields allows for reliable extraction by the models. The minor improvements observed with the Code Prompt compared to the NL Prompt on these fields can likely be attributed to its stricter format constraints.

For semi-structured fields, such as chief complaint, present illness history and past medical history, the F1 scores for all combinations remained high, ranging between 0.88 and 0.95, indicating good performance. Notably, the Code Prompt generally exhibited slight performance improvements over the NL Prompt for these types of fields, particularly for fields like “past history”. This suggests that code-formatted prompts can more effectively guide the model in capturing the key elements of semi-structured information.

Regarding unstructured fields, “preliminary diagnosis” achieved high extraction accuracy (F1 scores approximately 0.96-0.97) due to its relatively standardized nature. However, the “treatment recommendations” field presented a significant challenge in this evaluation. F1 scores were noticeably lower across all combinations, ranging from a minimum of 0.803 to a maximum of 0.847, and exhibited high variability. This is primarily attributed to the highly individualized and diverse phrasing of treatment recommendations, which poses challenges for consistent extraction by the models. Notably, the Code Prompt had the most pronounced impact on improving performance for this challenging field. Concurrently, the NL Model also demonstrated a superior capability compared to the Coder Model in handling this type of diverse language (e.g., achieving F1 scores of 0.847 vs. 0.810, respectively, under the Code Prompt).

Overall, the Code Prompt contributes to enhancing extraction accuracy and consistency, particularly when dealing with semi-structured fields and those with flexible phrasing. The NL Model holds an advantage in understanding and processing language-intensive information. However, the extraction of fields such as "treatment recommendations" remains a challenge. This highlights the need for future research to explore approaches like integrating medical knowledge bases to address the difficulties posed by the diversity in physician expression styles, thereby further enhancing model performance on complex medical information extraction tasks.

\subsection{Scenario Verification}

\begin{figure}[ht]
    \centering
    \includegraphics[width=\linewidth]{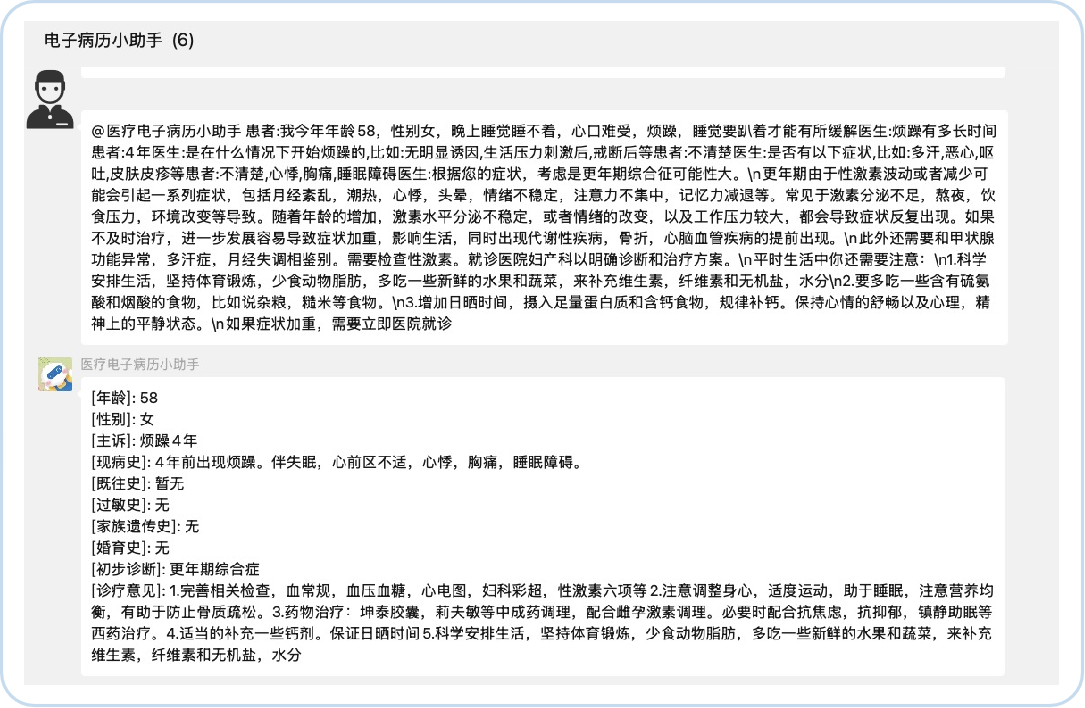} 
    \caption{\textbf{WeChat platform EMR generation assistant interaction example.}
}
    \label{fig:Scenario_Verification}
\end{figure}

To validate the practical value of EMRModeL (NL model + Code prompt), we successfully deploy it into two key applications: first, an EMR generation assistant based on the WeChat platform, and second, an EMR generation support tool integrated into the hospital information system. On the WeChat platform, the model is encapsulated as a convenient medical record extraction assistant. Healthcare professionals and related staff simply input the raw medical text in a conversational format. The system then automatically extracts core information such as medical history, chief complaint, and diagnostic opinions, and generates a structured electronic medical record. Figure \ref{fig:Scenario_Verification} clearly illustrates the entire interaction process between the user and the assistant on the WeChat interface: after the user submits a query via text, the assistant accurately understands the intent and quickly returns a standardized summary of medical information. This application scenario not only highlights the model's excellent capabilities in natural language understanding and information extraction but also demonstrates its good adaptability on mobile platforms, providing an efficient and intelligent auxiliary tool for clinical work.

\subsection{Discussion}

Based on the above experimental analysis, we believe that the prompt design is important, the LoRA fine-tuning has obvious advantages, and the model combination is feasible. The specific analysis is as follows:

\begin{itemize}
    \item \textbf{Importance of Prompt Design}

Experimental data indicate that a well-designed code-style prompt can significantly guide the model to better understand task requirements, resulting in more uniform and complete structured medical records. Compared to models using only natural language style prompts, models using code prompts show a notable advantage in F1 scores. For example, the F1 score of the coder model + code prompt is 85.4\%, higher than the 85.0\% of the coder model + nl prompt. Furthermore, the F1 score for the natural language large model + coder-style prompt fine-tuning is 88.1\%, higher than the 87.7\% of the natural language large model + nl prompt. Additionally, the models demonstrate more stable performance in Standard Deviation of F1 Score. This phenomenon indicates that structured prompts not only optimize the model's ability to capture key information but also improve the stability of the output format. 
    \item \textbf{Advantages of LoRA Fine-Tuning}

The use of LoRA fine-tuning technology significantly reduces the computational cost and parameter update requirements during the fine-tuning process, while enabling the model to adapt more quickly to the specific context of the medical domain. Experimental results show that the performance of the fine-tuned natural language large model in fine-grained information extraction tasks is significantly improved. When no fine-tuning was applied and the medical record summarization was directly performed on the base model, the pure natural language model achieved an F1 score of 49.0\%, while the coder model achieved 38.6\%. After LoRA fine-tuning, both the NL large model and the coder large model achieve F1 scores exceeding 80\%. This significant improvement fully demonstrates that LoRA technology helps enhance the model's ability to extract detailed information and effectively improves the accuracy and stability of model outputs. This provides strong theoretical and practical support for achieving efficient and low-cost model adaptation in medical scenarios.
    \item \textbf{Model Combination Effect}

Introducing a code-style prompt into the natural language large model strategy effectively combines the large model's broad capabilities in language understanding with the advantages of structured prompts specifically designed for information extraction tasks. Experimental results show that this combination strategy achieved the highest F1 score (88.1\%) and the lowest Standard Deviation of F1 Score on the test set, outperforming traditional coder model combinations. This indicates that combining structured prompt fine-tuning strategies can effectively address the lack of structured feature understanding in pure natural language large models for information extraction tasks, further improving the accuracy and consistency of key information extraction and output in medical consultation dialogues. 
\end{itemize}

Overall, in the task of medical consultation dialogue information extraction, well-designed code-style prompts and the LoRA fine-tuning technique play a significant role in enhancing model performance. While this study has achieved promising results, several challenges and areas for improvement remain. First, the diversity of linguistic expression and the presence of noise in real-world medical consultation dialogues pose challenges. Enhancing the model's robustness to such edge cases is a key direction for future research. Second, while the current fine-tuning strategy enhances domain adaptation, further exploration is warranted on how to better integrate medical knowledge bases with model outputs to improve interpretability and clinical trustworthiness. Lastly, in clinical applications, the model's response speed and the interpretability of its results are equally crucial. Future work will consider incorporating multimodal information and explainability mechanisms to better serve real-world diagnostic requirements.

\section{Conclusion }

The proposed EMRModel, integrating LoRA fine-tuning technology and code-style prompt design, effectively optimized the task of generating structured medical records from medical consultation dialogues. By leveraging large pre-trained language models and the low-rank adaptation technique, we significantly enhanced the efficiency and accuracy of medical text processing tasks. First, we constructed a high-quality medical dataset that not only covers multiple disease types and consultation scenarios but also includes extensive medical terminology, providing a solid foundation for training and evaluation. Second, through the introduction of LoRA fine-tuning and code-style prompt design, we successfully enhanced the model's sensitivity to domain-specific medical semantics and achieved significant performance improvements in the structured medical record generation task. Finally, we developed a benchmark framework for medical consultation record extraction, providing a standardized evaluation tool for future NLP research in the medical domain. Experimental results demonstrated that EMRModel holds significant advantages over traditional methods in the structured medical record generation task, particularly showcasing excellent accuracy and stability in terms of F1 score and the standard deviation of F1 scores across samples. 

Looking ahead, the research outcomes of EMRModel can be extended to multi-hospital and multi-department medical record generation tasks. Further exploration can involve integrating medical knowledge bases and retrieval-augmented generation mechanisms to enhance the medical accuracy and trustworthiness of the generated records. Simultaneously, we plan to further explore the application of this model in downstream tasks such as intelligent diagnostic assistance, automated medical record generation, and personalized treatment plan recommendations.

\section*{Acknowledgments}

This research is funded by the Anhui Postdoctoral Scientific Research Program Foundation (No. 2024A768) and the Major Science and Technology Project of Anhui Province (No. 202103a07020011).

\section*{Acknowledgments}

The authors declare no competing interests.

\bibliography{sn-bibliography}

\end{document}